\documentclass[runningheads]{llncs}
\usepackage[T1]{fontenc}
\usepackage{graphicx}
\usepackage{amsmath}
\usepackage{amssymb}
\usepackage{tikz}
\usepackage{xspace}
\usepackage{algorithm}
\usepackage{algorithmic}
\usepackage{bm}
\usepackage{bbding}
\usepackage[numbers]{natbib}
\begin{document}
\title{Leveraging Structural Constraints for Diffusion-based Neural TSP Solvers}
\titlerunning{PCI}
%\author{Anonymous submission}

\author{Micka\"el Basson\inst{1,2,3,4}\Envelope \and
Philippe Preux\inst{1,2,3,4}} %\and

%Third Author\inst{3}\orcidID{2222--3333-4444-5555}
%
%\authorrunning{Anonymous}
\authorrunning{M. Basson, Ph. Preux}

%\institute{Anonymous}
\institute{Université de Lille, France \and 
CNRS, France \and
Inria, France \and 
UMR 9189-CRIStAL, Lille, France \\
\email{\{mickael.basson, philippe.preux\}@inria.fr}}
%Springer Heidelberg, Tiergartenstr. 17, 69121 Heidelberg, Germany

\maketitle              % typeset the header of the contribution
\begin{abstract}
Neural combinatorial optimization has recently achieved str\-ong results on the Euclidean Traveling Salesman Problem (TSP) using generative models such as diffusion and consistency models. State-of-the-art approaches like FT2T combine fast consistency-based prediction with gradient-based inference time refinement. However, gradient search often incurs significant computational overhead and may not align with the discrete structure of feasible solutions. We introduce Projected Consistency Inference (PCI), a plug-and-play, retraining-free alternative that replaces gradient refinement with structure-aware projections: PCI decodes valid Hamiltonian tours from the consistency model output and applies a lightweight local search (e.g., 2-opt). PCI achieves an average optimality gap (OG) of 0.17\% on TSP with 500 cities, and 0.31\% on TSP with 1000 cities, outperforming FT2T best settings (OG 0.22\% and 0.36\%, respectively) while reducing the inference time up to 30 to 40\%. PCI also exhibits lower variance and memory usage, and can surpass classical heuristics such as LKH3 in rapid solution generation. Our results demonstrate that structure-aware inference time operations provide a practical and principled path for neural TSP solvers, complementing training time objectives.
\end{abstract}
\section{Introduction}
Combinatorial Optimization (CO) problems, such as the Traveling Salesman Problem (TSP), remain central to logistics, scheduling, network design, %and
chip layout and many other areas. Yet these problems are NP-hard, making it challenging to solve large instances with exact methods~\citep{helsgaun2017lkh3}. Classical heuristics, e.g., LKH3 for the TSP, are remarkably strong but require expert hand-crafting and careful parameter tuning~\citep{helsgaun2017lkh3}. 

%\pp{Solving a CO problem instance may be modeled as a supervised learning task. For a particular CO problem, we may assume we have a training set made of a set of instances of this problem, each labeled with its optimal solution. Once trained, the model would be able to generalize to yet unseen instances. This is precisely the aim of ``Neural CO'', where ``neural'' refers to the fact that the model is a neural network. Today, on the TSP, a neural model is trained on a large collection of random instances, e.g. up to one million instances; then, the model may solve very quickly other instances of the same problem, of any size. To liiustrate this point, in this paper, we report performances that competes with LKH3 on the TSPlib instances, being even faster than LKH3 at resolution time.}
Solving a CO problem instance may be modeled as a supervised learning task. For a particular CO problem, we may assume we have a training set made of instances of this problem, each labeled with its optimal solution. Once trained, the model would be able to generalize to yet unseen instances. ``Neural CO'', where ``neural'' refers to the fact that the model is a neural network, leverage the generalization ability of properly trained neural networks to this aim.
Neural CO has emerged to reduce manual design while leveraging data. Early approaches based on pointer networks and then on attention-based transformers used sequence models to \emph{construct} tours \citep{vinyals2015ptrnet,kool2019attention}. However, their performances were not matching those of dedicated heuristics well-known in the CO community. More recently, diffusion and consistency-based generative paradigms have become competitive for CO by modeling instance-conditioned solution distributions and enabling powerful inference time refinement~\citep{sun2023difusco,li2023t2tco, basson2024ideq, li2025fastt2t}. In particular, DIFUSCO cast CO as structured denoising on graph adjacency matrices and delivered strong results on the TSP and the MIS~\citep{sun2023difusco}. T2TCO introduced a gradient-based inference time search~\citep{li2023t2tco} and FT2T replaced multi-step diffusion sampling with a consistency predictor for one step or few steps generation, again paired with inference time gradient search steps leading to massive speed-up and improved solution quality~\citep{li2025fastt2t}. 

While inference time optimization is a powerful lever, two issues persist. First, iterative denoising/search can be time-consuming, particularly at scale and when augmented with gradient search steps ~\citep{sun2023difusco,li2023t2tco}. Second, gradient search steps operate in a relaxed latent continuous space which geometry may not align with the discrete and combinatorial nature of CO problems. %discrete feasibility or with the local-improvement that classical heuristics used for post-processing exploit.
IDEQ recently showed that injecting structural bias --- most notably, biasing toward 2-opt basins and leveraging constraints of the TSP solution manifold --- significantly improves the quality of solutions, and the generalization capability of the trained model~\citep{basson2024ideq}. %\pp{At the time of writing, IDEQ is the state of the art of neural approaches for the TSP, both on artificial instances and on the instances of the TSPlib.}

In this paper, we introduce \emph{Projected Consistency Inference (PCI)} that adapts IDEQ structural advantage at inference time to the consistency setting. We introduce a plug-and-play replacement for FT2T gradient search. For a given instance, PCI generates a solution in 2 steps: 
(i) a \emph{feasibility projection} from the continuous latent space to the set of Hamiltonian tours, %decodes valid discrete solutions (Hamiltonian tours) from the consistency predictor output;
(ii) a \emph{local search projection} %performs an improvement
to obtain a local optimum for a small computation cost (e.g., a 2-opt or 3-opt local optimum for TSP).
PCI requires no retraining. Instead it leverages the  previously trained models. Interestingly, PCI aligns with a broader trend in generative modeling that improves the quality of solution generation by refining the inference steps ~\citep{ma2025inference,heek2024multistep}.
PCI is the new state-of-the-art diffusion-based constructive neural method to solve TSP problems in terms of both speed and performance.

\paragraph{Contributions.}
This paper presents two main contributions.
(1) We formalize \emph{Projection-Enhanced Consistency Inference} (PCI) and instantiate it for the TSP as %a drop-in
an alternative to FT2T gradient search, requiring no additional training~\citep{li2025fastt2t}. 
(2) Empirically we show that PCI achieves significant quality and time improvement over FT2T gradient search on the TSP and is even able to outperform LKH3 on new instances drawn from the training distribution solved with a small computation time constraint.

We argue that structure-aware inference time operations are a principled and practical path for neural CO solvers, complementing training time objectives.

%%%%%%%%%%%%%%%%%%%%%%%%%%%%%%%%%%%%%%%%%%%%%%%%%%%%%%%%%%%%%%%%%%%%%%%%%%%%%%%%%%%%%%
%%%%%%%%%%%%%%%%%%%%%%%%%%%%%%%%%%%%%%%%%%%%%%%%%%%%%%%%%%%%%%%%%%%%%%%%%%%%%%%%%%%%%%

\section{Background and Related Work}

In this section, after a very brief recap on the TSP, we present the main components PCI rely on, that is diffusion models and their descendants, consistency models. Aside a brief presentation of these models, we present their use in the case of the resolution of the TSP. This section is completed by a brief overview of other neural CO methods though they do not compete with diffusion-based neural CO methods.

\subsection{About the TSP}

We consider the basic and usual definition of the Traveling Salesman Problem (see e.g. \cite{johnsonBiblicalArticle}). An instance of the TSP is defined in the Euclidean plane by a set of $N$ cities. $N$ is also known as the ``size'' of the instance. We will denote TSP-$N$ a TSP instance of size $N$.
The goal is to find a tour of minimal length that goes through all the cities.
The TSP is a NP-hard problem.
Today, the state-of-the-art exact solver Concorde \cite{concorde} solves an instance of the TSP of size $N=10^3$ in about 10 minutes on a laptop. For larger instances, one has to use heuristics that have no formal guarantee to find an optimal tour.
The quality of a tour may be measured by its ``optimality gap'', defined by $\frac{\mbox{length of the tour} - \mbox{optimal tour length}}{\mbox{optimal tour length}}$: it is a non-negative real number equal to 0 only for an optimal tour\footnote{In the case of large instances we use the best solution provided by heuristics as a proxy for the optimal solution. In this case, it is possible that the optimality gap is negative. Indeed, if the heuristics gives a non-optimal solution any solution between optimality and the heuristics one has a negative optimality gap. This has a very limited impact in practice.}.

An optimal tour goes once and only once per city: it is a Hamiltonian tour. Consequently, the search space may be reduced to the set of Hamiltonian tours. 
Given a Hamiltonian tour, a very simple heuristic known as ``2-opt'' may decrease its length. Given a Hamiltonian tour, a ``2-change'' consists in removing a pair of edges of the tour and reconstructing another tour (a single tour may be reconstructed). In 2D, when a pair of edges cross each other, applying a 2-change uncrosses the edges: this decreases the length of the tour. 2-opt considers all pairs of crossing edges and disentangles the pair that reduces the most the length of the resulting tour. 2-opt is very simple and rather effective in optimizing a tour. In practice, it is customary to pipeline a first heuristic that produces a Hamiltonian tour, and post-process it by iterating 2-opt to obtain a disentangled tour. There exists other such local optimizations, like the Lin-Kernighan operator (lk-opt) \cite{lk} and its more recent version LKH3 \cite{helsgaun2017lkh3} that are more complex in terms of algorithm, though very efficient and effective in practice.

\subsection{Diffusion for Combinatorial Optimization}
\paragraph{Diffusion Models}
constitute a class of generative algorithm that sample from complex unknown distributions such as distributions of images or graphs. They do so by mimicking the inverse of a noising process. Indeed, by progressively adding carefully selected random perturbations to clean samples drawn from the target distribution we obtain pure noise after a sequence of steps. %, pure noise.
The careful choice of the noisy perturbations ensures that the resulting noise is an easy to sample from distribution such as a Gaussian distribution or a  Bernoulli distribution. This is called the forward process. The goal of a diffusion model is to learn the reverse process, hence called the backward process: starting from a noise sample, it is denoised through a sequence of (backward) steps to obtain % iteratively denoise it to
a clean image or graph. Diffusion models use the generalization capabilities of neural networks to learn this process in a supervised manner with the training samples generated from the forward process. %: predicting the clean target from the (partially) noisy samples.
%From a noisy data, it predicts the original (clean) data.
%A diffusion model is just that: starting from an object drawn from an easy to sample from distribution (a uniform distribution), it (re-)constructs a complex object. 
%This is a complex process to learn, but advances in neural networks in the last 2 decades led to this sort of achievements. 
%In the following, we will leverage this idea to complex objects that are TSP instances and the associated adjacency matrices of their optimal solution. %Just to give the taste of it, we will encode a tour with a ``stochastic'' adjacency matrix: by this, we mean that this is a sort of adjacency matrix in which, instead of 0/1 entries, the entries of the matrix specify a quantity that is interpreted as the probability that a vertex (city) is connected to an other. Starting with a random adjacency matrix in which each entry is uniformly drawn from $[0, 1]$ (noisy solution), the diffusion process progressively denoise it so that in the end, we obtain an adjacency matrix that can be exploited to provide a Hamiltonian tour. Generating such a random adjacency matrix is easy, and denoising it will provide the complex object we are looking for that is difficult to sample, a Hamiltonian tour. Please, refer to \cite{yang2022diffusion} for a comprehensive introduction on diffusion models.}

This diffusion process may be expressed in discrete time or in continuous time.
%\pp{In a more technical way, }
%Diffusion Models (DMs) learn to generate data by reversing a fixed corruption (``noising'') process. 
In discrete-time, the Denoising Diffusion Probabilistic Model (DDPM)~\cite{ho2020denoising} starts from clean data $\mathbf{x}_0$ and sequentially adds Gaussian noise through the forward process (denoted as $q(\mathbf{x}_t \mid \mathbf{x}_{t-1})$ where $\mathbf{x}_{t}$ is the generated sample at timestep $t$):% gradually adds Gaussian noise to $\mathbf{x}_0$,
\begin{align}
q(\mathbf{x}_t \mid \mathbf{x}_{t-1}) &= \mathcal{N}\!\big(\sqrt{1-\beta_t}\,\mathbf{x}_{t-1},\,\beta_t \mathbf{I}\big), \quad t=1,\dots,T,\\
\mathbf{x}_t &= \sqrt{\bar{\alpha}_t}\,\mathbf{x}_0 + \sqrt{1-\bar{\alpha}_t}\,\boldsymbol{\epsilon},\quad \boldsymbol{\epsilon}\sim\mathcal{N}(\mathbf{0},\mathbf{I}),
\end{align}
where $\beta_t$ is the noise schedule, $\alpha_t=1-\beta_t$, $\bar\alpha_t=\prod_{s=1}^t \alpha_s$, and $\mathcal{N}(\mathbf{\mu},\mathbf{C})$ is the multivariate normal distribution of mean $\mathbf{\mu}$ and covariance $\mathbf{C}$. The backward process is parameterized as $p_\theta(\mathbf{x}_{t-1}\!\mid\!\mathbf{x}_t)=\mathcal{N}(\boldsymbol{\mu}_\theta(\mathbf{x}_t,t),\sigma_t^2\mathbf{I})$ and trained by predicting the noise $\boldsymbol{\epsilon}_\theta$ with the simplified objective\footnote{Other parameterizations exist, this one is the most frequently used.}:
\begin{align}
\mathcal{L}_{\text{simple}} = \mathbb{E}_{t,\mathbf{x}_0,\boldsymbol{\epsilon}}\big[\;\lVert \boldsymbol{\epsilon} - \boldsymbol{\epsilon}_\theta\!\big(\sqrt{\bar\alpha_t}\mathbf{x}_0 + \sqrt{1-\bar\alpha_t}\boldsymbol{\epsilon},\,t\big)\rVert_2^2\big],
\end{align}
where $\mathbf{\theta}$ denotes the set of parameters of the neural network and is used as a subscript to indicate a predictor function parameterized by such a network.

Continuous-time score-based formulations cast diffusion as a stochastic differential equation (SDE) $d\mathbf{x} = f(\mathbf{x},t)\,dt + g(t)\,d\mathbf{w}$ whose reverse-time SDE depends on the score $\nabla_{\mathbf{x}}\log p_t(\mathbf{x})$,
\begin{align}
d\mathbf{x} = \big[f(\mathbf{x},t) - g(t)^2 \nabla_{\mathbf{x}}\log p_t(\mathbf{x})\big]\,dt + g(t)\,d\bar{\mathbf{w}},
\end{align}
with an equivalent probability flow ordinary differential equation (PFODE) enabling deterministic sampling and likelihoods,
\begin{align}
\frac{d\mathbf{x}}{dt} = f(\mathbf{x},t) - \tfrac{1}{2} g(t)^2 \nabla_{\mathbf{x}}\log p_t(\mathbf{x}). 
\label{PFODE}
\end{align}
%These viewpoints unify DDPMs and score-based models and enable improved sampling schemes and likelihood estimation~\cite{song2021score}. 
Practical refinements—e.g., learned reverse variances and cosine noise schedules—further improve sample quality and reduce the number of steps~\cite{nichol2021improved}.

Discrete state space diffusion models \cite{austin2021d3pm} extend DDPMs to categorical variables $\mathbf{x}_t \in \{1,\dots,K\}^D$. 
The forward process is a Markov chain with transition matrices $Q_t \in [0,1]^{K\times K}$% applied independently per dimension:

\begin{align}
q(\mathbf{x}_t \mid \mathbf{x}_{t-1}) = \mathrm{Cat}(x_t ; p= x_{t-1}Q_t).
\end{align}

The marginal distribution at step $t$ can be expressed as:
\begin{align}
q(\mathbf{x}_t \mid \mathbf{x}_0) = \mathrm{Cat}(x_t ; p= x_0\bar{Q}_t) ,
\qquad \bar{Q}_t = Q_t Q_{t-1}\dots Q_1,
\end{align}
where $\bar{Q}_t$ is the cumulative transition matrix. 
The reverse process predicts categorical distributions with the following parametrization, where a neural network (with weights denoted $\mathbf{\theta}$) is trained to predict $\tilde{p_{\theta}}(\tilde{x_0}\mid x_t)$: 
\begin{align}
p_\theta(\mathbf{x}_{t-1}\mid \mathbf{x}_t) \propto \sum_{\tilde{x}_0} q(x_{t-1}, x_t\mid \tilde{x}_0) \tilde{p_{\theta}}(\tilde{x_0}\mid x_t).
\end{align}
%and training minimizes the discrete ELBO:
%\begin{align}
%\mathcal{L}_{\text{D3PM}} = \mathbb{E}\Big[-\log p_\theta(\mathbf{x}_0\mid \mathbf{x}_1) + \sum_{t=2}^T \mathrm{KL}\big(q(\mathbf{x}_{t-1}\mid \mathbf{x}_t,\mathbf{x}_0)\,\|\,p_\theta(\mathbf{x}_{t-1}\mid \mathbf{x}_t)\big)\Big].
%\end{align}
This model supports structured transitions (e.g., absorbing states, lattice kernels) and achieves strong likelihoods on text and image benchmarks.

%\pp{Applied to the TSP, the diffusion process acts on stochastic adjacency matrices, that is a matrix which entries $(i, j)$ is proportional to the probability that there exists an edge between city $i$ and city $j$. We can also see such a matrix as a heatmap. the denoising model is trained on (instance, optimal tour adjacency matrix) pairs so that starting from an arbitrary (noisy, and easy to sample) matrix $\mathbf{x}_T$, the denoising process produces $\mathbf{x}_0$ which is a (stochastic) adjacency matrix of an optimal tour. However, the denoising process can not by itself really produce an adjacency matrix of a Hamiltonian tour. $\mathbf{x}_0$ contains a small number of entries that are significantly larger than the others that points towards the edges of a tour, so that $\mathbf{x}_0$ can be easily transformed into a Hamiltonian tour (je ne sais pas s'il faut dire ce qui suit ici car c'est IDEQ qui le fait, pas DIFUSCO.) which can be further optimized by a local search algorithm, such as iterating 2-opt.} 

\paragraph{DIFUSCO} introduced diffusion models on graphs to solve CO problems. % on many CO problems are naturally defined on a graph.}% graph-based diffusion for CO.% formulating binary solution variables and denoising schedules on graphs with effective inference schemes. 
The solution to the COP can be seen as the adjacency matrix of a graph. This (discrete) adjacency matrix $A$ is relaxed to a real $[0,1]^{N\times N}$ matrix ($N$ being the number of nodes) with a probabilistic interpretation: the component $a_{ij}$ of $A$ is the probability that the edge between nodes $i$ and $j$ is present in the solution. This matrix is commonly referred to as a heatmap. DIFUSCO casts TSP and MIS as an instance conditioned heatmap generation problem. For the TSP, the predicted heatmap is then projected onto a Hamiltonian tour that is further optimized by a local search algorithm, such as iterating 2-opt.
%DIFUSCO set strong baselines on the TSP and the MIS but requires multi-step sampling, creating a quality–time tension at inference~\citep{sun2023difusco}. 
\paragraph{Training-to-Testing: T2TCO.}
Following DIFUSCO, T2TCO proposed the train\-ing-to-testing paradigm: learn an instance-conditioned distribution (via diffusion) and then conduct gra\-dient-based search at inference time to exploit objective gradients for instance-specific improvement~\citep{li2023t2tco}. %This narrows the gap between training distributions and test instances but retains iterative sampling plus inference time search overhead~\citep{li2023t2tco}. 
\paragraph{Structure-Aware Training: IDEQ.}\label{sec:structure-aware-IDEQ}
Following DIFUSCO and T2TCO, IDEQ significantly improves diffusion-based TSP solvers by leveraging the constrained TSP solution manifold and taking into account the local search in the training objective. It does so by projecting the candidate solution at each time step of the backward process to the manifold of locally optimal Hamiltonian tours. %Instead of applying the probabilistic relaxation to the entire generative process IDEQ does so at each step individually. 
IDEQ also sets a new training objective: instead of one optimal solution, IDEQ is searching for the set of tours that maps to the same local optimum. So, instead of searching for a single point, IDEQ searches for a basin of attraction: this introduces redundancy in the training objective and increases the entropy of the training set. This significantly enhances the performance of IDEQ with respect to DIFUSCO and T2TCO on TSPslib and synthetic TSPs~\citep{basson2024ideq}.

%\pp{Je ne sais pas si c'est facile à faire, mais je trouverais intéressant pour que le lecteur (et moi ;-) comprenne bien d'illustrer ce processus de débruitage sur une toute petite instance comme ulysses16 : une figure avec $\mathbf{x}_T$, débruitage progressif jusqu'au heatmap $\mathbf{x}_0$, le tour hamiltonien qui en résulte puis le 2-opt optimum. Je pense que l'on a tout à gagner à ce que le lecteur comprenne le processus, même si les maths sont dures pour lui. Par ailleurs, que vaut $T$ ? combien d'étapes de débruitage sont-elles faites~?}

\subsection{Consistency Models for CO}
\paragraph{Consistency models (CM)} aim to reduce the computational cost of the multistep sampling done in diffusion models by learning a mapping that is invariant across noise levels. 
%To that end, consistency models (CMs)~\cite{song2023consistency} learn a function $f_\theta(\mathbf{x}_t, t)$ that maps a noisy input $\mathbf{x}_t$ at noise level $t \in [0,T]$ directly to a clean sample (or an intermediate state) while satisfying a \emph{consistency condition}. 
%Let $\mathbf{x}_t$ be obtained by adding noise to a data sample $\mathbf{x}_0$ according to a predefined schedule. 
%The consistency condition requires that for any $s, t \in [\epsilon, T]$\footnote{\pp{$\epsilon$ is a small value close to 0 meant to avoid numerical instabilities of the algorithm.}}:
%\begin{align}
%f_\theta(\mathbf{x}_t, t) \approx f_\theta(\mathbf{x}_s, s),
%\end{align}
%where $\mathbf{x}_s$ and $\mathbf{x}_t$ belong to the same probability flow ODE trajectory. 
%To that end, consistency models (CMs)~\cite{song2023consistency} learn a function $f_\theta$ that maps a noisy input $\mathbf{x}_t$ at noise level $t \in [0,T]$ directly to a clean sample: $f_\theta(\mathbf{x}_t, t) =\mathbf{x}_0$. This consistency function is constant over a given probability flow ODE trajectory. This ensures that the model output is invariant across noise levels, enabling denoising to be performed in a single step or in a very few steps.%one step or few steps generation.
The core idea of a Consistency Model is to learn a function $\bm{f}_\theta: (\mathbf{x}_t, t) \mapsto \mathbf{x}_\epsilon$ that maps any point $\mathbf{x}_t$ on the PFODE trajectory (\ref{PFODE}) to the starting point of the trajectory $\mathbf{x}_\epsilon$ (which approximates the data $\mathbf{x}_0$). That is, for any trajectory $\{\mathbf{x}_t\}_{t \in [\epsilon, T]}$ that satisfies the PFODE, the consistency function must satisfy the \textbf{self-consistency property}:
\begin{equation}
    \bm{f}_\theta(\mathbf{x}_t, t) = \bm{f}_\theta(\mathbf{x}_{t'}, t') \quad \forall t, t' \in [\epsilon, T].
\end{equation}
Essentially, the output of the model is invariant to the time $t$ along the integration path.
To ensure that the output is valid, the model must satisfy the boundary condition at the minimum time step $\epsilon$:
\begin{equation}
    \bm{f}_\theta(\mathbf{x}_\epsilon, \epsilon) = \mathbf{x}_\epsilon.
\end{equation}
In practice, this is usually enforced via skip-connection parameterization using a deep neural network $N_\theta$:
\begin{equation}
    \bm{f}_\theta(\mathbf{x}, t) = c_{\text{skip}}(t)\mathbf{x} + c_{\text{out}}(t)N_\theta(\mathbf{x}, t),
\end{equation}
%where $c_{\text{skip}}(t)=1$ and $c_{\text{out}}(t)=0$.
where $c_{\text{skip}}(\epsilon)=1$ and $c_{\text{out}}(\epsilon)=0$.
%\subsection{The Consistency Loss}
%The general loss function minimizes the difference between the model's prediction at the current time step $t_{n+1}$ and a target estimate derived from the next time step $t_n$ (closer to data).

%Given a metric $d(\cdot, \cdot)$ (e.g., $\ell_2$, LPIPS) and a data sample $\mathbf{x}$, we generate $\mathbf{x}_{t_{n+1}} \sim \mathcal{N}(\mathbf{x}, t_{n+1}^2 %\mathbf{I})$. The loss is:
%\begin{equation}
%    \mathcal{L}(\theta, \theta^-) = \mathbb{E}_{n, \mathbf{x}, \mathbf{z}}\left[ \lambda(t_n) \cdot d\left( \bm{f}_\theta(\mathbf{x}_{t_{n+1}}, t_{n+1}), \: \bm{f}_{\theta^-}(\hat{\mathbf{x}}_{t_n}^{\phi}, t_n) \right) \right]
%\end{equation}
%where:
%\begin{itemize}
%    \item $\theta$ are the online weights being optimized.
%    \item $\theta^-$ are the target weights (usually an Exponential Moving Average (EMA) of $\theta$).
%    \item $\hat{\mathbf{x}}_{t_n}^{\phi}$ is an estimate of $\mathbf{x}_{t_n}$ obtained from $\mathbf{x}_{t_{n+1}}$.
%\end{itemize}

%\textbf{Consistency training} enforces this condition by minimizing:
%\begin{align}
%$\mathcal{L}_{\text{consistency}} = \mathbb{E}_{s,t,\mathbf{x}_t}\Big[d\big(f_\theta(\mathbf{x}_t, t), f_\theta(\mathbf{x}_s, s)\big)\Big]$,
%\end{align}
%where $d(\cdot,\cdot)$ is typically an $\ell_2$ or perceptual distance. 
Two main training strategies are used: consistency distillation from a teacher diffusion model, or direct consistency training. 
These approaches allow CM to achieve competitive sample quality with drastically fewer steps than diffusion models, hence for a much reduced computational cost.%, while supporting zero-shot editing tasks such as inpainting and super-resolution.

\paragraph{FT2T} accelerates the T2TCO pipeline by replacing iterative diffusion sampling with a consistency predictor that maps noisy states directly to high-quality data, optionally in a few steps. It uses an inference time consistency-based gradient search to further explore the learned solution space~\citep{li2025fastt2t}. In our experiments, we find that such a gradient refinement adds computational overhead while providing limited quality gains, motivating a structure-aware, retraining-free alternative.

Our work adopts the structural insights from IDEQ at inference time, composing them with FT2T one step (or a few steps) consistency to obtain PCI, without any retraining needed.

%\paragraph{Beyond our focus (brief).}
%Neural constructive/improvement solvers (e.g., hierarchical constructive for realistic TSP; GELD for extreme scales) report competitive or SOTA performance on various TSP regimes~\citep{goh2024hierarchical,xiao2025geld}. In the broader generative literature, \emph{consistency models} and \emph{multistep consistency} provide foundations for one/few-step generation and stability analyses~\citep{song2023consistency,heek2024multistep}. Recent work in vision validates \emph{inference-time} search/verification as an orthogonal lever to improve sample quality without retraining~\citep{ma2025inference}.
\subsection{Other Neural CO Approaches}

To avoid the costly generation of the supervised training dataset, other approaches leverage unsupervised learning (like UTSP \cite{utsp}) or reinforcement learning (RL). RL-based CO solvers are either autoregressive (incremental construction of the solution), or generating the full solution in one-shot. The former category includes BQ-NCO \cite{bqnco}, POMO \cite{pomo} and SymNCO \cite{symnco} while the latter category includes DIMES \cite{dimes}. DIMES employs a meta-learning framework and a continuous parameterization of the solution space which enables a stable REINFORCE-based training.
In BQ-NCO the state space of the Markov Decision Problem associated with the problem is simplified by bisimulation quotienting. 
POMO builds on the attention-based model of \cite{KoolAM} but leverages symmetries by data augmentation and stabilizes REINFORCE training by using multiple initiations. SymNCO also leverages symmetries by introducing regularization in REINFORCE and by learning invariant representation for pre-identified symmetries. SymNCO and POMO do not scale well beyond 200 cities for the TSP. ICAM \cite {ICAM} builds upon POMO, modifying the attention mechanism and the reinforcement learning training scheme to significantly improve performance on larger TSP instances (up to $10^3$ cities). 

Another line of work aims at building solutions by splitting the problem into simpler and smaller sub-problems and merging together the partial solutions. To the best of our knowledge GLOP \cite{glop} is the current SOTA for such methods. 

Aside pure constructive solvers other encouraging research areas have focused at improving generalization via knowledge distillation \cite{KD}, meta-learning \cite{Omni}, ensemble methods \cite{Gao2024} or local encoding \cite{Invit}. 

Finally, ensemble methods leverage the diversity of existing solvers to improve solution generation \cite{Gao2025, poppy}

%%%%%%%%%%%%%%%%%%%%%%%%%%%%%%%%%%%%%%%%%%%%%%%%%%%%%%%%%%%%%%%%%%%%%%%%%%%%%%%%%%%%%%
%%%%%%%%%%%%%%%%%%%%%%%%%%%%%%%%%%%%%%%%%%%%%%%%%%%%%%%%%%%%%%%%%%%%%%%%%%%%%%%%%%%%%%

\section{Projected Consistency Inference}
\label{sec:method}

\subsection{Projection Operators}
\begin{figure}[!ht]
    \centering
    \begin{tikzpicture}
        \node[anchor=south west, inner sep=0](image) at (0,0){
            \includegraphics[width=0.65\linewidth]{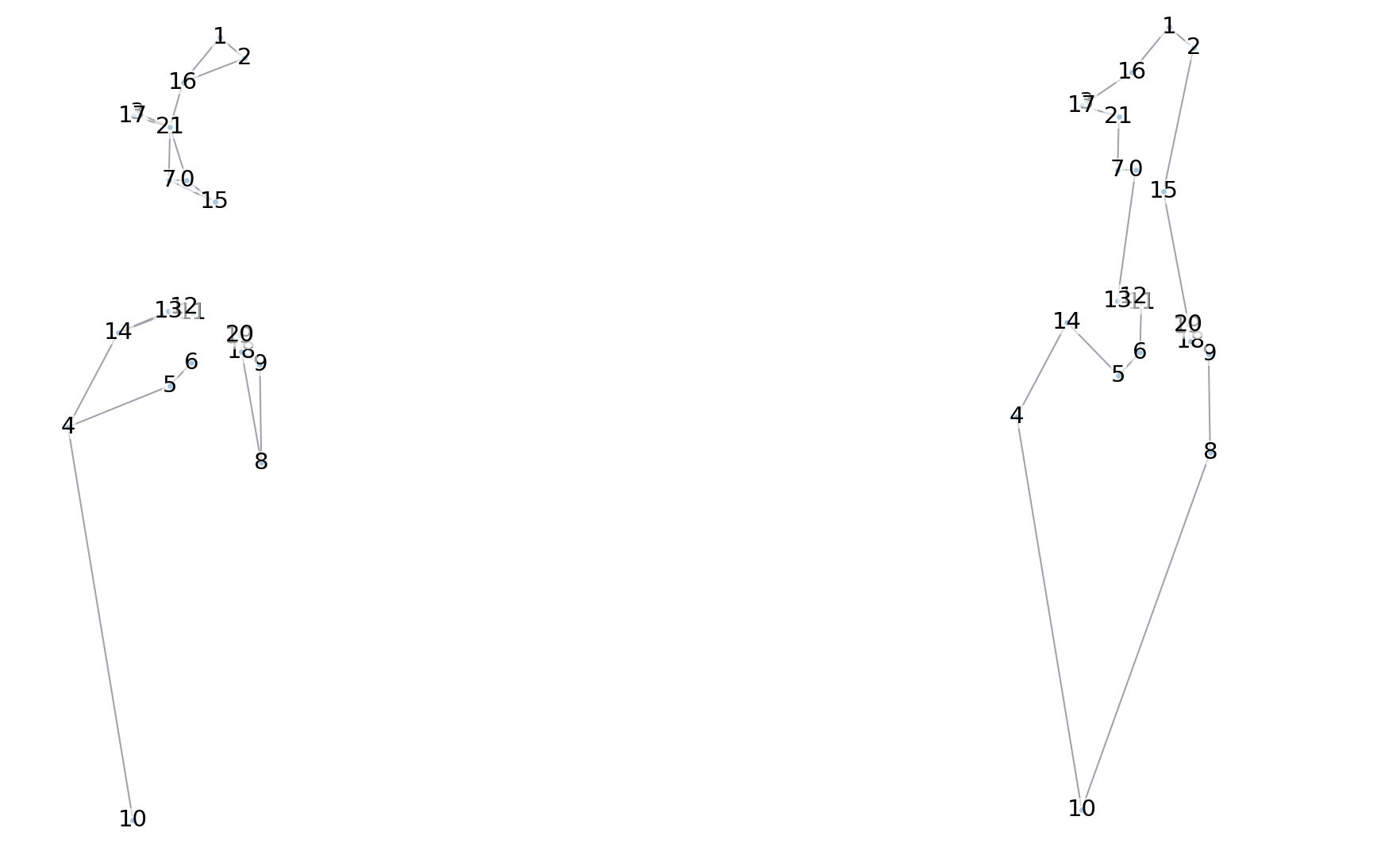} %, [...trim={0.5cm 1.2cm 0.5cm 1.2cm},clip]
            };
            \begin{scope}[x={(image.south east)},y={(image.north west)}]
            \draw[black, thick, ->] (0.3,0.5) -- (0.6,0.5);
            \node[above] at (0.45,0.5) {$P_\mathrm{Local}(P_\mathrm{Feas})$};
            \end{scope}
    \end{tikzpicture}
    \caption{PCI projection operations illustrated on the TSPlib instance Ulysses22. We use a very small instance for legibility purposes. The left graph is the diffusion-predicted $\hat{x}_0$. %On this real example,
      Before applying $P_\mathrm{Feas}$, the graph may be made of more than one component. Then, the graph on the right results from the application of the 2 projections, $P_\mathrm{Feas}$ and then $P_\mathrm{Local}$: it is a 2-opt locally optimal Hamiltonian tour.}%, the right graph is $P_\mathrm{Local}(P_\mathrm{Feas}(\hat{x}_0))$
    
    \label{fig:project}
\end{figure}

Build on earlier works, we propose PCI which composes two projections:
\begin{enumerate}
\item \textbf{Hamiltonian Tour Reconstruction ($P_\mathrm{Feas}$).} Given edge probabilities from $f_\theta$ (a heatmap), PCI maps it to a Hamiltonian tour. This is the same operator as applied at the end of the DIFUSCO inference pipeline to reconstruct a valid tour from the heatmap (iterative greedy edge insertion procedure based on the heatmap probabilities which excludes edges that do not produce a valid tour) \cite{sun2023difusco}.
\item \textbf{Local Search Operator ($P_\mathrm{Local}$).} On the Hamiltonian tour resulting from $P_\mathrm{Feas}$, PCI applies a local search routine (e.g., 2-opt) to obtain a locally optimal tour.
\end{enumerate}
This procedure is illustrated in figure \ref{fig:project} on a small instance of the TSPliB.

\subsection{PCI Algorithm}
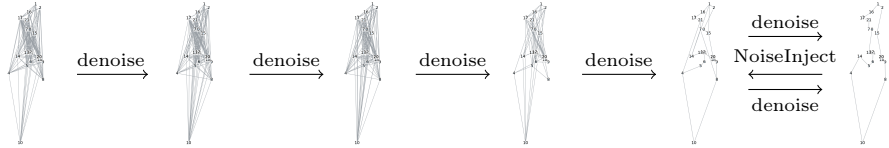
\begin{figure}[!ht]
    \centering
    \begin{tikzpicture}
        \node[anchor=south west, inner sep=0](image) at (0,0){
            \includegraphics[width=\linewidth]{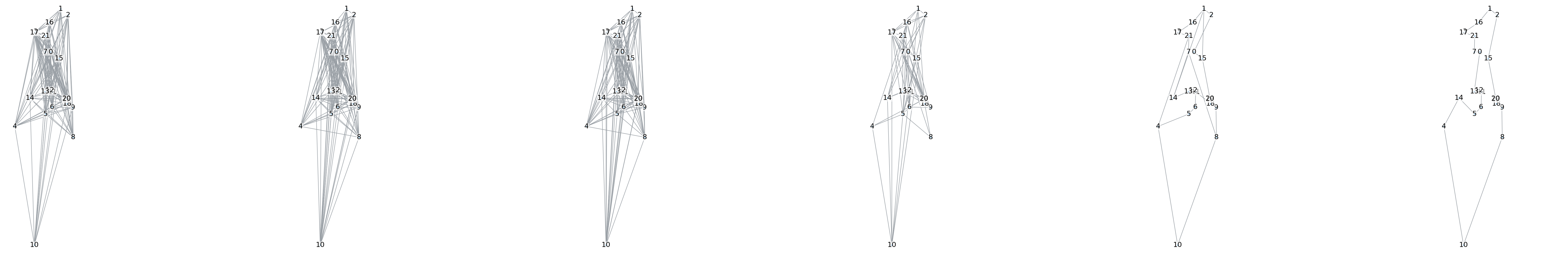} %, [...trim={0.5cm 1.2cm 0.5cm 1.2cm},clip]
            };
            \begin{scope}[x={(image.south east)},y={(image.north west)}]
            \draw[black, thin, ->] (0.0833,0.5) -- (0.16,0.5);
            \node[above] at (0.12,0.5) {\scriptsize{denoise}};
            \draw[black, thin, ->] (0.27,0.5) -- (0.35,0.5);
            \node[above] at (0.31,0.5) {\scriptsize{denoise}};
            \draw[black, thin, ->] (0.45,0.5) -- (0.53,0.5);
            \node[above] at (0.49,0.5) {\scriptsize{denoise}};
            \draw[black, thin, ->] (0.63,0.5) -- (0.71,0.5);
            \node[above] at (0.67,0.5) {\scriptsize{denoise}};
            \draw[black, thin, ->] (0.81,0.75) -- (0.89,0.75);
            \node[above] at (0.85,0.75) {\scriptsize{denoise}};
            \draw[black, thin, <-] (0.81,0.5) -- (0.89,0.5);
            \node[above] at (0.85,0.5) {\scriptsize{NoiseInject}};
            \draw[black, thin, ->] (0.81,0.4) -- (0.89,0.4);
            \node[below] at (0.85,0.4) {\scriptsize{denoise}};
            \end{scope}
    \end{tikzpicture}
    \caption{PCI denoising and re-noising illustrated on Ulysses22 TSPlib instance. The left graph is a random graph with edge probability $p=1/2$ and the right one is the predicted solution. The denoise operation is the joint consistency prediction and projections: $\text{denoise} \sim P_\mathrm{Local}(P_\mathrm{Feas}(f_{\theta}))$. We also illustrate the noise injection process. For conciseness we only depict once the partially noised graph (second to the right) and the re-denoised graph but on larger instances these usually differ leading to a solution improvement. There may also be more than one noise injection.}
    \label{fig:denoise}
\end{figure}

We interleave consistency prediction (same as original consistency model), projection, and re-noising. The re-noising is the same as used in the orginal consistency model, it is denoted here as $NoiseInject$\footnote{Using the same notations as above, $NoiseInject: x_0, t \mapsto x_t \sim Cat( x_t, p=x_0\bar{Q}_t$)}. This is illustrated in figure \ref{fig:denoise} and the full algorithm is given in Algorithm \ref{alg:pci}.%: 

%\[
%x_0^{(k)} \leftarrow P_\mathrm{Local}\big(P_\mathrm{Feas}(f_\theta(x_t^{(k)}, t^{(k)}, G))\big), \quad
%x_t^{(k+1)} \leftarrow \text{NoiseInject}(x_0^{(k)}, t^{(k+1)}).
%\]
%In the single step case ($K=1$), PCI reduces to decoding and projecting once. In the few steps case, this procedure is repeated for stability.

\begin{algorithm}[h]
\caption{Projected Consistency Inference (PCI)}
\label{alg:pci}
\begin{algorithmic}[1]
\REQUIRE Trained consistency model $f_\theta$, instance $G$, number of denoising steps $K$, noise schedule $\{t^{(k)}\}$, number of renoising steps $J$, renoise time $\tau$ 
\STATE Initialize $x_t^{(1)} \sim q(\cdot)$ // Backward process
\FOR{$k=1$ \TO $K$}
    \STATE $\hat{x}_0 \leftarrow f_\theta(x_t^{(k)}, t^{(k)}, G)$
    \STATE $x_0^{(k)} \leftarrow P_\mathrm{Local}(P_\mathrm{Feas}(\hat{x}_0))$
    \STATE $x_t^{(k+1)} \leftarrow \text{NoiseInject}(x_0^{(j)}, t^{k+1})$
\ENDFOR
\STATE Initialize $x_0^{(0)} \leftarrow x_0^{(K)}$ // Renoise steps
\FOR{$j=0$ \TO $J-1$}
    \STATE $x_{\tau} \leftarrow \text{NoiseInject}(x_0^{(j)}, \tau)$
    \STATE $\hat{x}_0 \leftarrow f_\theta(x_{\tau}, \tau, G)$
    \STATE $x_0^{(j+1)} \leftarrow P_\mathrm{Local}(P_\mathrm{Feas}(\hat{x}_0))$
\ENDFOR
\STATE \textbf{return} $\{x_0^{(j=0)}, ..., x_0^{(j=J)}\}$
\end{algorithmic}
\end{algorithm}

PCI eliminates gradient computations and uses deterministic projections with time-limited local optimization, reducing wall clock by approximately $40\%$ with respect to the FT2T gradient search under matched computational budgets. 

Crucially, PCI is \emph{ready out-of-the-box}: it requires no retraining, no hyperparameter tuning, and integrates seamlessly with existing FT2T pre-trained models.% (so-called ``checkpoints'' in their jargon).

%%%%%%%%%%%%%%%%%%%%%%%%%%%%%%%%%%%%%%%%%%%%%%%%%%%%%%%%%%%%%%%%%%%%%%%%%%%%%%%%%%%%%%
%%%%%%%%%%%%%%%%%%%%%%%%%%%%%%%%%%%%%%%%%%%%%%%%%%%%%%%%%%%%%%%%%%%%%%%%%%%%%%%%%%%%%%
\section{Experimental Study}

\subsection{Experimental Methodology}

Experiments were conducted on samples drawn from the FT2T training distributions that is made of uniformly distributed 2D Euclidean TSP with sizes $N=500$ and $N=1000$. The corresponding pre-trained neural models from FT2T were used; these models are available on the Internet on the FT2T GitHub repository \cite{fastt2t_github}. We also conducted experiments on samples from the TSPlib \cite{tsplib}, a highly renown benchmark in the TSP community. We used the instances of the TSPlib with size ranging from $10^2$ to $10^4$ cities with the model pretrained on 1000 cities instances. This tests both the ability to generalize beyond the training distribution and the training size. 
To be consistent with previous publications \cite{basson2024ideq, li2023t2tco, li2025fastt2t, sun2023difusco}, the running times being reported are times to solve 128 instances. However, due to the stochastic nature of the generated solution, we averaged the optimality gap and the time to solve 128 instances over $128^2=16384$ instances to get more accurate performance mean and variability estimates.
Statistical test of superiority to compare optimality gaps were performed using Welch test. For the TSPlib we measured the mean and variance of the optimality gap over 48 replicas of the dataset, with different random seeds, to ensure enough statistical power.

\paragraph{On Sampling Size.}
The official FT2T implementation \cite{fastt2t_github} does not perform a pure gradient refinement loop; instead, at each step it concatenates the gradient-guided output with the naive (non gradient-guided) one, effectively doubling the candidate pool per iteration. This design introduces a parallel sampling effect beyond the theoretical gradient only update. In our definition the sampling size is the effective number of parallel candidate solutions being processed simultaneously. So a sampling size $S$ from FT2T (for which we keep the notation $S$ to be consistent with the published paper and released code) has to be compared to PCI with sampling size $2S$. Such comparison matches the effective candidate count under similar compute budgets. This reflects the true inference time diversity leveraged by FT2T and PCI. 

\paragraph{Hardware}

Experiments involving GPU were all carried on the same cluster of 8 GPU A100 40GB. A single GPU was used for each experiment except for the TSP-1000 with sampling size 4 and above where the 8 GPUs were used in parallel. Execution times were linear in the number of GPUs and are reported in the table as GPU-min (i.e.\@ 1 min on 8 GPUs = 8 GPU minutes). 

LKH3 running time was measured on a Ryzen 9 5900X CPU with 24 cores. 
All running times are measured based on the sequential processing of the set of instances to make comparison easier. 

\subsection{Experimental Results}

First, we compared FT2T and PCI on random instances generated from Euclidean 2D TSP with 500 cities (results in table \ref{main_res500}) and 1000 cities (results in table \ref{main_res1000}). We see that PCI outperforms FT2T both in terms of compute speed and of performance. PCI also exhibits lower variance. The average optimality gap differences were all significant with $p<.0001$ (Welch test).

\begin{table*}[h]
  \centering
  \caption{Comparison of optimality gaps (OG) -reported as mean (standard deviation)- and running times, averaged over 128 instances, on large scale 2D uniformly distributed Euclidean TSP instances. xS denotes a parallel sampling of size S. Note that as explained in the main text Fast-T2T effectively doubles the sampling size so fast-T2T x2S compares with PCI xS and there is no comparison for PCI x1.} %Between brackets: standard deviation of the OG, calculated over 16384 instances}
  \begin{tabular}{|l|c|c||c|c|}
    \hline
    & \multicolumn{4}{|c|}{TSP-500}   \\
    & \multicolumn{2}{|c||}{1 step, 1 renoise} & \multicolumn{2}{|c|}{5 steps, 5 renoises} \\
    Method & OG. & time & OG. & time   \\
    \hline
    PCI x1                 &0.90\% (0.47) & 0.20mn & 0.49\% (0.28) & 0.78mn \\
%    Consistency model x1    &0.91 & 48mn & 0.46 & 2h15 \\
    \hline
    PCI x2                  &0.73\% (0.38) & 0.32mn & 0.34\% (0.21)& 1.3mn \\
%    Consistency model x2    &0.73 & 1h32 & 0.32 & 4h14 \\
    Fast-T2T x1             &0.74\% (0.49) & 0.55mn & 0.40 \%(0.59) & 1.9mn \\
    \hline
    PCI x8                 &0.48\% (0.27)& 1.0mn & 0.17\% (0.13)& 4.0mn \\
%    Consistency model x8    &0.48 & 4h55 & 0.16 & 13h13 \\
    Fast-T2T x4             &0.54\% (0.73) & 1.8mn & 0.22\% (0.46) & 5.8mn \\ 
    \hline
  \end{tabular}
  \label{main_res500}
\end{table*}

\begin{table*}[h]
  \centering
    \caption{Comparison of optimality gaps (OG) -reported as mean (standard deviation)- and running times, averaged over 128 instances, on large scale 2D uniformly distributed Euclidean TSP instances. xS denotes a parallel sampling of size S. Note that as explained in the main text Fast-T2T effectively doubles the sampling size so fast-T2T x2S compares with PCI xS and there is no comparison for PCI x1.}
  \begin{tabular}{|l|c|c||c|c|}
    \hline
    & \multicolumn{4}{|c|}{TSP-1000}   \\
    & \multicolumn{2}{|c||}{1 step, 1 renoise} & \multicolumn{2}{|c|}{5 steps, 5 renoises} \\
    Method & OG. & time & OG. & time   \\
    \hline
    PCI x1                 &1.13\% (0.37) & 0.69mn & 0.70\% (0.26) & 2.4mn \\
%    Consistency model x1    &1.13 & 3h07 & 0.64 & 8h52 \\
    \hline
    PCI x2                 &0.94\% (0.31) & 1.1mn  & 0.53\% (0.21) & 4.3mn \\
%    Consistency model x2    &0.95 & 5h32 & 0.49 & 16h30 \\
    Fast-T2T x1             &1.0\% (0.34) & 2.1mn & 0.57\% (0.23) & 7.0mn \\
    \hline
    PCI x8                 &0.67\% (0.23) & 4.0mn & 0.31\% (0.14) & 16.5mn \\
%    Consistency model x8    &0.67 & 2h42 x8 & 0.28 & 7h58 x8\\
    Fast-T2T x4             &0.75\% (0.26) & 7.8mn & 0.36\% (0.16) & 25.5mn \\ 
    \hline
  \end{tabular}
  \label{main_res1000}
\end{table*}

Then we compared FT2T and PCI on instances of the TSPlib. We used the model trained on TSP-1000 instances to solve instances of size ranging from 100 to $10^4$ cities. Results are displayed in table \ref{main_resTSPlib}. PCI outperformed FT2T both in terms of performance and running time. The average optimality gap differences were all significant with $p<.0001$ (Welch test). We also noticed the impact of the gradient search on the memory: GPU RAM requirement is significantly lower with PCI. This results in the ability of PCI to solve larger instances on the same hardware. 

\begin{table*}[h]
  \centering
    \caption{Comparison of optimality gaps (OG) -reported as mean (standard deviation)- and running times on TSPlib instances whose size ranges from $10^2$ to $10^4$ cities. xS denotes a parallel sampling of size S. Note that as explained in the main text Fast-T2T effectively doubles the sampling size hence the absence of data to compare with PCIx1. The running time is reported to solve once the entire dataset (76 instances).}
  \begin{tabular}{|l|c|c||c|c|}
    \hline
    & \multicolumn{4}{|c|}{TSPlib 100-10,000}   \\
    & \multicolumn{2}{|c||}{1 step, 1 renoise} & \multicolumn{2}{|c|}{5 steps, 5 renoises} \\
    Method & OG. & time & OG. & time   \\
    \hline
    PCI x1                 &1.95 \%(1.62) & 1.4mn & 1.37\% (1.39) & 4.4mn \\
    \hline
    PCI x2                 &1.58\% (1.43) & 2.4mn & 1.15 \%(1.36) & 8.4mn \\
    Fast-T2T x1             &1.71 \%(1.53) & 4.1mn & 1.25\% (1.43) & 13.8 mn \\
    \hline
    PCI x8                 &1.13 \%(1.35) & 8.3~mn & 0.83\% (1.25) & 32.2~mn \\
    Fast-T2T x4             & \multicolumn{2}{|c||}{ Out of memory }  & \multicolumn{2}{c|}{ Out of memory } \\ 
    \hline
  \end{tabular}
  \label{main_resTSPlib}
\end{table*}

These results show that PCI is able to generalize to instances with sizes and/or distributions differing from the ones used for the training of the underlying model.
Outperforming FT2T, PCI establishes the new state-of-the-art performance of neural CO methods.

\subsection{Comparison With LKH-3}

\begin{figure}[!ht]
    \centering
    %\includegraphics[width=\linewidth]{var\XYZ.png}
    %\includegraphics[width=\linewidth]{varT2TCO.png}
    %\caption{Distribution of predicted versus ground truth tour length obtained over 32 repeats of 64 TSP-1000 instances with \XYZ (top) and T2TCO (bottom).}
    \includegraphics[width=.9\linewidth]{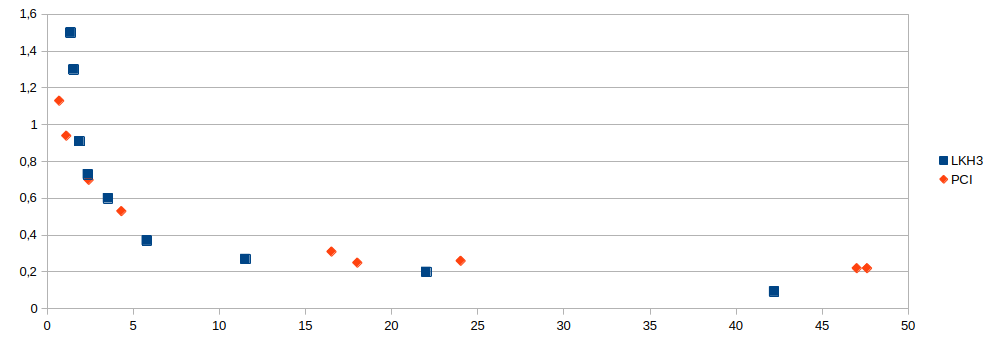} %, [...trim={0.5cm 1.2cm 0.5cm 1.2cm},clip]
    \caption{Total time (x-axis, in minutes) versus mean optimality gap (y-axis, in percentage) performance of PCI and LKH3 on 128 TSP-1000 random instances.}
    \label{fig:LKH3-1K}
\end{figure}

\begin{figure}[!ht]
    \centering
    %\includegraphics[width=\linewidth]{var\XYZ.png}
    %\includegraphics[width=\linewidth]{varT2TCO.png}
    %\caption{Distribution of predicted versus ground truth tour length obtained over 32 repeats of 64 TSP-1000 instances with \XYZ (top) and T2TCO (bottom).}
    \includegraphics[width=0.55\linewidth]{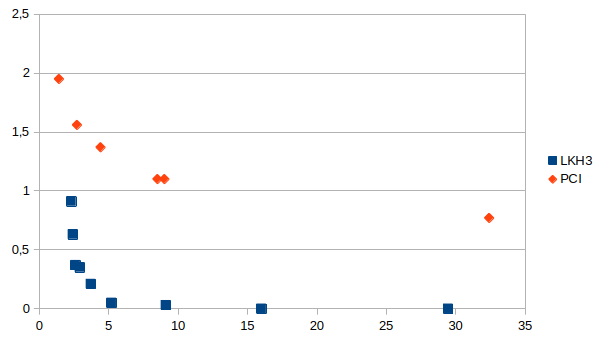}
    \caption{Total time (x-axis, in minutes) versus mean optimality gap (y-axis, in percentage) performance of PCI and LKH3 on the 76 TSPlib instances with size ranging from$10^2$ to $10^4$ cities.}
    \label{fig:LKH3-TSPlib}
\end{figure}

We compared the time/per\-for\-man\-ce tradeoffs of PCI and LKH3. The time/per\-for\-man\-ce tradeoff for PCI is based on the various settings given in table \ref{main_res1000} i.e.\@ sampling size and number of steps. For LKH3 this tradeoff is based on changing the number of runs ($10^2, 10^3, 10^4$)  and the max number of trials per run ($5, 10, 20$). For both parameters the central value is the default one. 
Fig.\@ \ref{fig:LKH3-1K} reports the results for instances drawn from the training distribution (Euclidean 2D uniform TSP-1000). Fig.\@ \ref{fig:LKH3-TSPlib} reports the same figures for the instances of the TSPlib.
%We show results for instances drawn from the training distribution (TSP-1000) and for instances of the TSPlib.


On these two plots, we note the remarkable performance of PCI on random instances: on these instances, PCI matches LKH3 time/performance trade-off. On TSPlib instances, for the same time budget, we note that PCI is generating tours that are not as good as those returned by LKH3, about 1\% off. However, for a specific instance (fl3795), PCI consistently outperformed LKH3. %, and even less within 30 minutes of computing time. 
It is well-known that the performance of various algorithms is not the same on random instances and on real-world instances: the latter are more structured, a feature that LKH3 is designed to exploit, whereas PCI is not able to do alike. This is crucially related to the training of underlying neural models.%: instead of relying on training instances that are solely drawn uniformly at random, we should draw them form a distribution of instances that are more similar to real instances or coming from a mix of diverse distributions.

Overall, PCI is the current state-of-the-art method in diffusion-based neural CO, but also comes very close to the performances of the state-of-the-art methods on the TSP when evaluated on distributions that match the training ones.

%%%%%%%%%%%%%%%%%%%%%%%%%%%%%%%%%%%%%%%%%%%%%%%%%%%%%%%%%%%%%%%%%%%%%%%%%%%%%%%%%%%%%%
%%%%%%%%%%%%%%%%%%%%%%%%%%%%%%%%%%%%%%%%%%%%%%%%%%%%%%%%%%%%%%%%%%%%%%%%%%%%%%%%%%%%%%

\section{Discussion}

We have demonstrated experimentally that PCI is an efficient and fast alternative to gradient search to improve inference time solution generation. It is worth noting that these benefits come out of the box without any retraining or fine-tuning the neural model: using publicly released, already pre-trained, models (in this work: \cite{fastt2t_github}); hence at no further training cost. %Indeed, we use models available on the Internet, trained by the FT2T authors.}

The gain in inference time is perfectly understood as PCI requires less computational steps; for the gain in solution quality we can  hypothesize the following: (1) FT2T motivates gradient-based refinement by arguing that following the objective gradient in a relaxed latent space should improve solution quality. However, the gradient is computed in a continuous relaxation of a discrete combinatorial space. This relaxation is highly non-convex and disconnected; small steps in the relaxed space do not guarantee any improvement in the discrete objective after decoding. (2) The gradient search uses an energy-based model of the tour length as a surrogate objective for the unknown ground truth optimal solution. Associated gradient refinement only explores a narrow neighborhood around each initial sample. This does not guarantee any improvement after local search. This contrasts with projection-based strategies such as the ones used in PCI which are not local.

Our empirical results confirm this: PCI, which enforces feasibility and leverages structure-aware local moves, outperforms gradient refinement under lower compute budgets.

\section{Conclusion and future work}

In this paper, we have presented PCI, our new neural CO solver, and demonstrated it on the TSP. We compared the experimental performance of PCI with the state-of-the-art diffusion-based neural network based method FT2T, as well as the LKH3 TSP dedicated heuristic. This comparison shows that PCI very significantly improves the state-of-the-art of diffusion-based neural methods, and though not outperforming LKH3, PCI comes close to its performance.
This research is applicable to all types of diffusion application where structural constraints can be formalized and enforced e.g.\@ through projections. This includes other combinatorial optimization problems, such as Vehicle Routing Problems.
Compared to LKH-3, our results show that while this neural method can match or even outperform (on short time scale and training distribution) decade-long handcrafted heuristics, it suffers from out-of-training distribution performance drop as shown on the TSPlib dataset.
Improving generalizability is one of our future lines of work. Our models are trained using uniformly random instances, and it is well-known that algorithms do not behave on random TSP instances as they do on real-world instances. This prompts the investigation of particular structured random instances to train such models (either to re-train them from scratch, or fine-tune existing models).
Similar to IDEQ, to constrain the diffusion to predict a solution on the space of locally optimal Hamiltonian tours, we used projections. Noticeably, the Hamiltonian tour projection uses a greedy insertion procedure. There are other projections of increasing complexity like beam-search or Monte-Carlo Tree Search that may lead to better results. 
The projection itself may not be the most effective approach to ensure constraint satisfaction; exploring alternative strategies is another line of future research.  

%\section*{Acknowledgments}
%Micka\"el Basson is also a full-time employee of Lilly France detached to work on this project. 
%We acknowledge the Région Hauts-de-France CPER project CornelIA. Experiments presented in this paper were carried out using the Grid’5000 testbed (https://www.grid5000.fr). 
% add minister of research 
%This work was granted access to the HPC resources of IDRIS under the allocation 2025-AD011015857R1 made by GENCI.
%Both authors acknowledge the Scool research group for an outstanding working environment.

\bibliographystyle{splncs04}
\bibliography{bibliography}

\end{document}